\documentclass{article}

\usepackage[english]{babel}

\usepackage[letterpaper,top=2cm,bottom=2cm,left=3cm,right=3cm,marginparwidth=1.75cm]{geometry}
\usepackage{todonotes}
\usepackage[bottom]{footmisc}
\usepackage{booktabs} 
\usepackage{enumitem}
\usepackage{amsmath}
\usepackage{amsfonts} 
\usepackage{graphicx}
\usepackage{subcaption}
\usepackage{listings} 
\usepackage{hyperref}
\usepackage{tcolorbox} 
\usepackage{mdframed} 

\usepackage{color}  
\definecolor{dullgreen}{rgb}{0.4, 0.6, 0.4}
\definecolor{paleYellow}{RGB}{255, 255, 220}

\usepackage{color}
\definecolor{codegreen}{rgb}{0,0.6,0}
\definecolor{codegray}{rgb}{0.5,0.5,0.5}
\definecolor{codepurple}{rgb}{0.58,0,0.82}
\definecolor{backcolour}{rgb}{0.95,0.95,0.92}

\definecolor{humanColor}{RGB}{220, 235, 255}
\definecolor{chatgptColor}{RGB}{235, 255, 220}

\newcommand{\humanLogo}{\raisebox{-0.1\height}{\includegraphics[width=12pt]{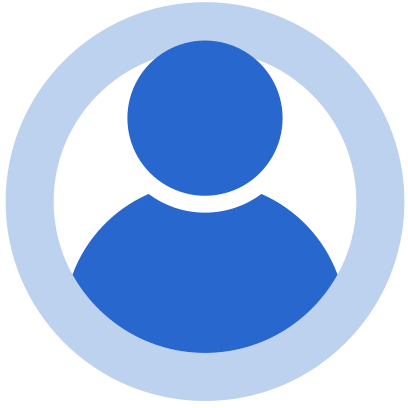}}} 
\newcommand{\chatgptLogo}{\includegraphics[width=12pt]{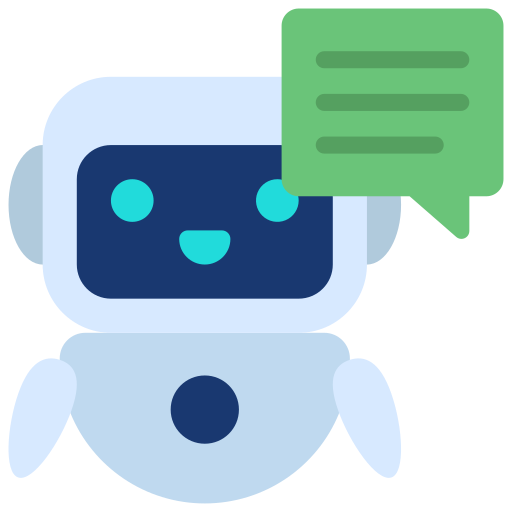}} 
\newcommand{\bgLogo}{\includegraphics[width=12pt]{chatbot.png}} 

\newmdenv[%
  leftmargin=0.1cm,
  rightmargin=0.2cm,
  innertopmargin=0.1cm,
  innerbottommargin=0.2cm,
  linewidth=0pt,
  linecolor=black,
  roundcorner=0.1cm,
  backgroundcolor=humanColor,
  font=\small\sffamily,
  skipbelow=-0.0cm, 
]{humanBox}

\definecolor{lightGrey}{gray}{0.9}

\newmdenv[%
  leftmargin=0.1cm,
  rightmargin=0.2cm,
  innertopmargin=0.1cm,
  innerbottommargin=0.2cm,
  linewidth=0pt,
  linecolor=black,
  roundcorner=0.1cm,
  backgroundcolor=paleYellow,
  font=\small\sffamily,
  skipbelow=-0.0cm,
]{bgBox}

\newmdenv[%
  leftmargin=0.1cm,
  rightmargin=0.2cm,
  innertopmargin=0.1cm,
  innerbottommargin=0.2cm,
  linewidth=0pt,
  linecolor=black,
  roundcorner=0.1cm,
  backgroundcolor=chatgptColor,
  font=\small\sffamily,
  skipbelow=-0.0cm, 
]{chatgptBox}

\newmdenv[%
  leftmargin=0.5cm,
  rightmargin=0.5cm,
  innertopmargin=0pt,
  innerbottommargin=0pt,
  linewidth=0.5pt, 
  linecolor=white,
  font=\small\sffamily,
  skipbelow=0pt,
  skipabove=0pt,
]{outerBox}
\mdfdefinestyle{myboxstyle}{
  linecolor=black, 
  linewidth=0.5pt, 
  frametitleaboveskip=8pt, 
  frametitlebelowskip=0pt, 
  innertopmargin=8pt, 
  innerbottommargin=8pt, 
  skipabove=0pt,
  skipbelow=0pt,
  nobreak=false, 
}

\newmdenv[style=myboxstyle]{mybox} 

\title{A Human-on-the-Loop Optimization Autoformalism Approach for Sustainability}
\author{Ming Jin\footnote{Corresponding author} , Bilgehan Sel, Fnu Hardeep, Wotao Yin}

\begin{document}
\maketitle

\begin{abstract}

This paper outlines a natural conversational approach to solving  personalized energy-related problems using large language models (LLMs). We focus on customizable optimization problems that necessitate repeated solving with slight variations in modeling and are user-specific, hence posing a challenge to devising a one-size-fits-all model.
We put forward a strategy that augments an LLM with an optimization solver, enhancing its proficiency in understanding and responding to user specifications and preferences while providing nonlinear reasoning capabilities.
Our approach pioneers the novel concept of \emph{human-guided optimization autoformalism}, translating a natural language task specification automatically into an optimization instance. This enables LLMs to analyze, explain, and tackle a variety of instance-specific energy-related problems, pushing beyond the limits of current prompt-based techniques.

Our research encompasses various commonplace tasks in the energy sector, from electric vehicle charging and Heating, Ventilation, and Air Conditioning (HVAC) control to long-term planning problems such as cost-benefit evaluations for installing rooftop solar photovoltaics (PVs) or heat pumps. This pilot study marks an essential stride towards the \emph{context-based formulation of optimization using LLMs}, with the potential to democratize optimization processes. As a result, stakeholders are empowered to optimize their energy consumption, promoting sustainable energy practices customized to personal needs and preferences.
\end{abstract}

\section{Introduction}

The escalating demand for energy, alongside the imperative to curb carbon emissions, has spurred a surge in interest in cultivating sustainable energy practices \cite{tyson2022americans}.  Substantial advancements have been made in the application of advanced computation such as optimization for Electric Vehicle (EV) charging \cite{arif2021review}, energy storage control \cite{khezri2022optimal}, renewable energy system investment \cite{yang2018battery}, smart building operation \cite{drgovna2020all}, and demand side management \cite{esther2016survey}, among others. However, the efficacy of these advancements is limited if they are not widely adopted and implemented by end-users \cite{rolnick2022tackling}.

Fundamentally, the adoption of optimization technology has faced challenges due to the high marginal costs associated with learning to model problems as optimizations, coding instances, and invoking functions for individual users. These costs establish barriers to entry, preventing many potential users from fully capitalizing on these advanced technologies. In particular, small businesses and individuals may struggle to access complex modeling and optimization tools due to a lack of necessary expertise or resources, rendering existing approaches insufficient \cite{currie2012opti}.

\begin{figure}[t]
  \centering
    \includegraphics[width=\textwidth]{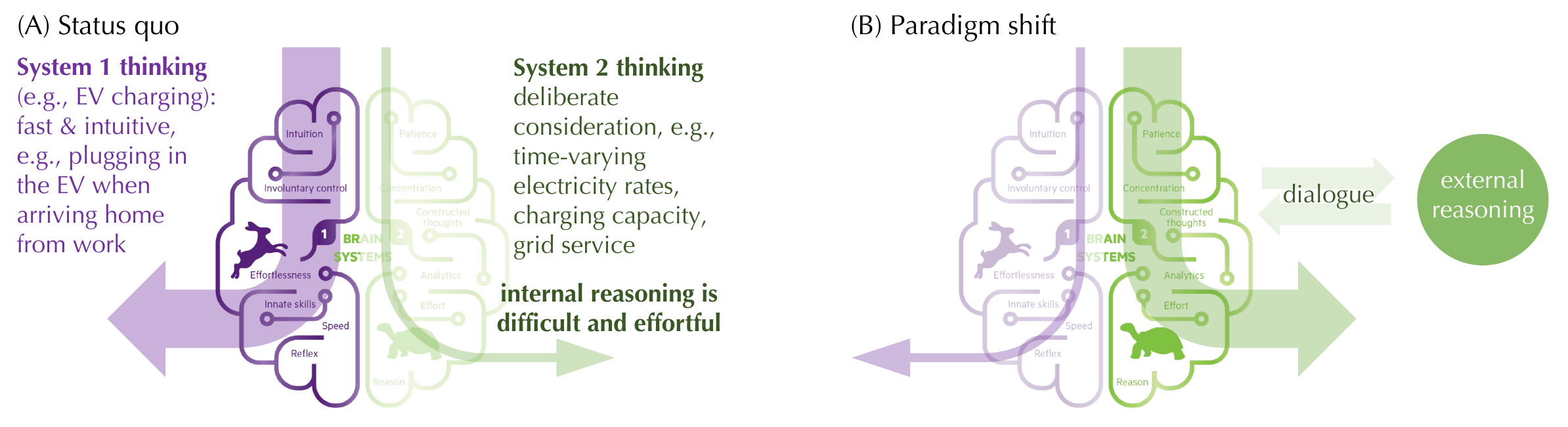}
  \caption{A comparison of decision-making approaches (e.g., EV charging). \textbf{(A)} Status Quo: Individuals typically engage in System 1 thinking, opting for immediate and convenient solutions. The effortful nature of System 2 thinking, which involves deliberate considerations, often leads to its disuse. \textbf{(B)} Paradigm Shift:  Our Energy Concierge activates System 2 thinking by delegating complex reasoning to an external engine and engaging users in easy-to-follow dialogues, promoting informed and sustainable practices.}
  \label{fig:overview}
\end{figure}

\paragraph{Dismantling barriers: How LLMs may democratize access to optimization solutions.} The advent of LLMs marks a paradigm shift in surmounting these obstacles \cite{kenton2019bert,brown2020language,chen2021evaluating,chatgpt}. In this paper, we demonstrate that by leveraging the power of LLMs, we can effectively bridge the gap between the high marginal costs of traditional optimization methods and the demand for personalized, accessible solutions. Although we do not aim to provide an industry-grade optimization formulation, our objective is to equip users with the tools necessary to make more informed decisions, ones that may involve formulating an optimization problem.  Our approach develops the potential of LLMs to streamline the modeling and optimization processes, enabling users to engage with the system through natural language instructions, instead of requiring extensive knowledge of programming or mathematical optimization techniques.

\paragraph{Energy Concierge: Offloading internal reasoning to an external engine.}

Nobel Laureate Kahneman describes two distinct systems of human thought: System 1, characterized by fast, automatic, and intuitive thinking, and System 2, which entails slower, more deliberate, and effortful reasoning \cite{kahneman2011thinking}. Adopting sustainable energy practices often necessitates System 2 thinking, such as closely monitoring energy consumption and making data-driven choices. However, System 1 thinking, being more automatic and less effortful, frequently dominates or supersedes System 2 (Fig. \ref{fig:overview}A). This paper introduces an \textbf{Energy Concierge} (EC) that aids users in managing their energy usage by offering personalized recommendations and real-time feedback.
\begin{quote}
The Energy Concierge facilitates user engagement with System 2 thinking by delivering tailored advice and immediate feedback on energy-related decisions (Fig. \ref{fig:overview}B). Through conversational interaction with the AI system, users can engage in more thoughtful, intentional, and reflective decision-making regarding their energy consumption, ultimately fostering increased energy efficiency and sustainable energy practices.
\end{quote}

\paragraph{Technical challenges and solutions.} While LLMs such as BERT \cite{kenton2019bert}, GPT-3 \cite{brown2020language}, Codex \cite{chen2021evaluating}, and OpenAI's ChatGPT \cite{chatgpt}  offer a natural substrate for conversational AI (a technology that enables machines to understand, interpret, and engage in natural conversations), it is essential to recognize that current LLMs have not excelled in tasks like arithmetic and logical reasoning \cite{rae2021scaling}. Recent works have introduced methods such as chain of thoughts \cite{weichain} (generalized by tree of thoughts \cite{yao2023tree} and algorithm of thoughts \cite{sel2023algorithm}), which prompts a series of intermediate reasoning steps, and autoformalism \cite{wuautoformalization}, which automatically translates natural language mathematics to formal specifications and proofs. However, many energy problems, like energy storage control and long-term planning for PV panel installation, demand complex decision-making. These problems differ from arithmetic reasoning, commonsense reasoning, and symbolic reasoning\footnote{Benchmark datasets for arithmetic reasoning include GSM8K \cite{cobbe2021training}, SVAMP \cite{patel2021nlp}, ASDiv \cite{miao2020diverse}, and MAWPS \cite{koncel2016mawps}, which use datasets such as math word problems with varying structures, diverse math word problems, and algebraic word problems. For commonsense reasoning, popular datasets include CSQA \cite{talmor2018commonsenseqa}, StrategyQA \cite{geva2021did}, and SayCan \cite{brohan2022can}, which cover a diverse range of commonsense reasoning types, such as answering questions about the complex world semantics and mapping natural language to robot actions. For symbolic reasoning, tasks include last letter concatenation and coin flip \cite{weichain}.}  in the following aspects: \emph{Complexity}, as they often involve numerous variables, constraints, and objectives with potentially nonlinear relationships between variables; and \emph{incomplete information}, as energy systems are influenced by various factors, such as user preferences, which may not be provided in the initial problem description.

In this research, we address each of these differences as follows. To manage complexity, we develop a procedure to map a problem description to code using the grammar of packages that support optimization formulation, and to iteratively formulate, debug, and execute the program. We also introduce external tooling to execute the written code and solve the formulated optimization using dedicated algorithms. Furthermore, we implement prompt engineering techniques to enable LLMs to accurately understand and respond to user-specific preferences. To address incomplete information, we leverage the reasoning capabilities of LLMs to identify key parameters and use a question-and-answer format in natural conversations to solicit this information from users. Additionally, we employ LLMs for auto-informalism to explain the solution to users. An overview of the EC framework is depicted in Fig. \ref{fig:ab}. This approach bears dual benefits: it not only guides the user through complex nonlinear reasoning tasks of energy saving planning but also offers an insightful explanation of the solution.

\begin{figure}[t]
  \centering
    \includegraphics[width=.7\textwidth]{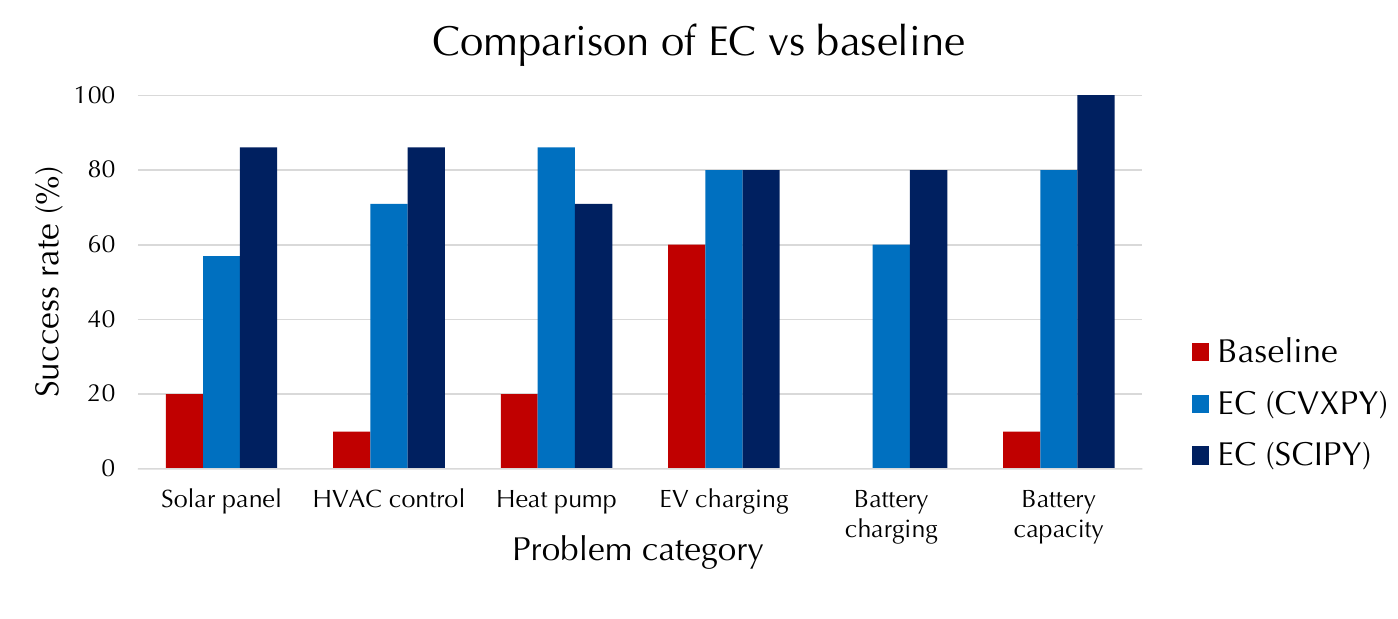}
  \caption{Comparison of the baseline method (simple prompting with GPT-4) and EC using either CVXPY or SCIPY as the code projection layer across various problem categories. Success rate is measured by the percentage of instances in which the provided answer aligns with user expectations, satisfies all given constraints, and is either close to or exactly optimal.}
  \label{fig:comparison}
\end{figure}

We further demonstrate the possibility of our proposed approach in solving a variety of tasks within the energy domain, ranging from EV charging, HVAC and battery control, to long-term planning problems such as cost-benefit evaluation of installing rooftop solar PVs, heat pumps, and battery sizing. In comparison to the simple prompting method, where the task description is directly presented to the LLM, our proposed EC framework improves the success rate, as illustrated in Fig. \ref{fig:comparison}. Additional details can be found in Sec. \ref{sec:experiments}. Although the optimizations formulated are not exceedingly sophisticated, we find that the LLM effectively addresses the problem to a considerable degree.
Throughout the remainder of the paper, unless specified otherwise, we refer to OpenAI's ChatGPT (GPT-4) when mentioning LLM.

The remaining sections of this paper are structured as follows. In Section \ref{sec:related_work}, we provide an overview of related work on conversational AI and optimization for sustainable energy practices. Section \ref{sec:method} describes our proposed framework in detail, including the design principles and the optimization semi-autoformalism approach. Section \ref{sec:experiments} presents experimental results that demonstrate the effectiveness of our approach in solving various energy-related tasks. We discuss the potential and future directions of conversational AI for energy sustainability and conclude in Section \ref{sec:conclude}.

\section{Related Work}\label{sec:related_work}
Our research resides at the confluence of large language models, autoformalization, and optimization.

\textbf{Autoformalization.} The recent success of LLMs has inspired numerous studies utilizing them in the field of mathematical autoformalization. This direction is concerned with the conversion of natural language into formal languages \cite{wu2022autoformalization, jiang2022thor}, such as Isabelle/HOL \cite{wenzel2008isabelle}. These formal languages operate under strict, definite rules. In the realm of convex optimization, a variety of solvers cater to different types, including linear or quadratic programs. In a manner analogous to formal languages in mathematics,  convex solvers accept problems in a so-called standard form \cite{agrawal2018rewriting}. The representation of this form can be challenging due to the lack of contextual correspondence. In our study, we use SCIPY \cite{virtanen2020scipy} and CVXPY \cite{diamond2016cvxpy} as intermediaries, bridging the gap between an intuitive method of formulating optimization problems and their transformation into the standard form. Mirroring the scenario in mathematical autoformalization in Isabelle, we observe that despite the scarcity of CVXPY examples online, large language models exhibit a surprisingly adept understanding of it, though they occasionally make syntactic errors. We correct these errors in our debugging process, leveraging feedback sourced directly from Python.

\textbf{Multi-step Problem Solving.} While LLMs excel at tasks solvable in a single step, they grapple with problems requiring multi-step coordination to yield a solution \cite{brown2020language, austin2021program}. Several strategies, such as prompting techniques \cite{kojima2022large}, in-context learning \cite{wei2022chain}, tree search \cite{yao2023tree}, and fine-tuning \cite{nye2021show}, have been employed to navigate the model through a problem progressively, demonstrating encouraging results. However, these techniques are less effective with tasks that demand nonlinear and iterative reasoning, such as the resolution of a standard optimization problem which typically necessitates thousands of iterative updates. Our research presents a novel viewpoint, employing optimization as a reasoning tool, thus enriching the spectrum of existing techniques (for an exhaustive review of LLM augmentation techniques, refer to the recent survey \cite{mialon2023augmented}).

\textbf{Revision with Feedback.} LLMs have exhibited their capacity to revise responses grounded in feedback across diverse contexts, such as social interactions \cite{bai2022constitutional, liu2023training}, question-answering \cite{gao2023rarr}, and code-debugging via unit tests \cite{olausson2023demystifying}. In our work, we exploit this potential by incorporating revisions based on direct debugging information supplied by programming languages or user requests within the operational framework. This methodology enables rapid rectification of semantic and syntactic inaccuracies and aids in clarifying ambiguities in inadequately described problems. Additionally, we estimate the success rate enhancement attributable to revision in our experiments.

\textbf{Conversational AI for Smart Buildings.} The resolution of optimization problems in the energy sector demands a profound understanding of commonsense reasoning, a familiarity with standard product feature parameters, and knowledge of typical objectives or requirements associated with specific problems \cite{saka2023conversational}. In the quest for energy efficiency and improved safety in smart homes, the majority of research has focused on the application of machine learning \cite{ghiaus2006experimental, yuan2018predictive, callaghan2001soft}, pattern recognition \cite{yu2016advances, kyriacou2017partitioning}, and expert systems \cite{panchalingam2021state}. Our method harnesses the inherent expertise of LLMs to frame optimization problems that can make informed decisions through user interactions. As far as we are aware, \emph{our study is the first to tackle non-trivial, energy-specific optimization problems with incomplete information purely through natural language specifications.}

\section{Energy Concierge Framework}
\label{sec:method}

\textbf{Design principles.} To ensure that conversational AI can effectively promote greater energy efficiency and sustainable energy practices, we propose a set of design principles. These include a modular pipeline, consisting of several components such as an LLM, code compiler, and user interface that can be used for a wide range of energy-related problems. Interactivity is also a key aspect, enabling users to engage with the system in a natural and conversational manner. Personalization is crucial, tailoring the system to each user's needs, preferences, and constraints. Real-time feedback is provided to users, offering timely information and allowing them to adjust their behavior accordingly. Lastly, explainability plays a significant role, involving the provision of clear explanations of the system's behavior, rationale, and limitations in order to improve user trust and acceptance of the system.

\begin{figure}[t]
  \centering
    \includegraphics[width=\textwidth]{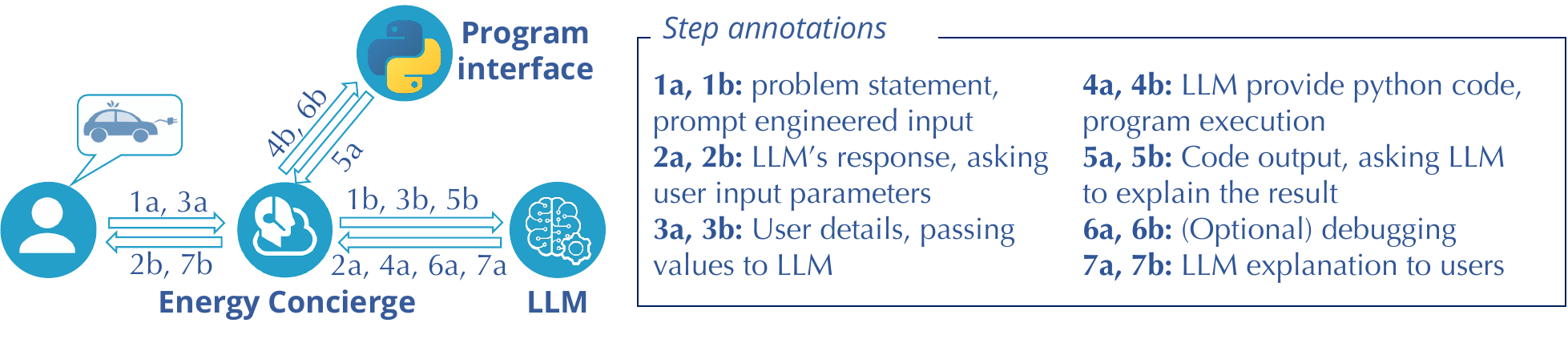}
  \caption{Energy Concierge framework. The user engages with an LLM through natural language queries and responses. The LLM identifies the necessary input parameters for optimization and generates Python code to address the problem. The program interface then executes the code and relays the solution back to the LLM, which subsequently provides a clear explanation to the user.}
  \label{fig:ab}
\end{figure}
In accordance with the design principles, we have developed the Energy Concierge framework, as depicted in Fig. \ref{fig:ab}. Using a simple EV charging example, the framework operates as follows:
\begin{enumerate}
    \item Initially, the user submits a natural language query or request concerning energy management (\textsc{Step} 1a), such as ``I need help optimizing my EV charging schedule to minimize costs.'' LLM (such as ChatGPT or GPT-4) processes the query, determining whether it should be treated as an optimization problem. If not (for instance, ``How can I reduce my energy bill?''), it generates a response in natural language.
    \item If the query necessitates optimization, the system's Prompt Engineered input  requests the LLM to identify specific input parameters needed to solve the problem (\textsc{Steps} 1b, 2a). Consequently, the LLM may ask the user questions like ``What is the charging capacity of your EV?'' and ``What are your preferred charging hours?'' (\textsc{Step} 2b).
    \item Upon receiving the user's responses (\textsc{Step} 3), the LLM formulates a Python code using the provided information (\textsc{Step} 4a). For instance, the LLM may generate code that models the EV charging schedule optimization problem based on the given charging capacity, charging hours, and electricity rates. This step, referred to as optimization autoformalism (Sec.~\ref{sec:optim}), automatically translates the natural language query into an optimization problem.
    \item The program interface then receives the generated code and executes it to solve the optimization problem (\textsc{Step} 4b). If necessary, the code may undergo iterations for debugging and refinement to achieve an accurate solution (\textsc{Steps} 6a, 6b).
    \item After obtaining the solution, the LLM explains the results, such as the optimal charging schedule and associated cost savings (\textsc{Step} 5b, 6a). The user is presented with this information in a clear and comprehensible manner, enabling them to make informed decisions regarding their EV charging practices (\textsc{Step} 7a, 7b).
\end{enumerate}

We provide technical details in the following subsections.

\subsection{Optimization autoformalism}
\label{sec:optim}

A general optimization problem can be written as:
\begin{equation}
\begin{aligned}
\text{minimize}_{x\in\mathbb{R}^p}\quad & f(x;\theta) \\
\text{subject to}\quad & g_i(x;\theta) \leq 0, \quad i=1,2,\dots,m\\
& h_j(x;\theta) = 0, \quad j=1,2,\dots,n,
\end{aligned}
\tag{$P(\theta)$}\label{opt-P}
\end{equation}
where we seek to minimize the objective function $f(x;\theta)$ subject to a set of inequality constraints $g_i(x;\theta) \leq 0$, and equality constraints $h_j(x;\theta) = 0$. Here, $x\in\mathbb{R}^p$ is the decision variable and $\theta$ is the collection of hyperparameters that defines the optimization instance, including objective and constraint functions; in other words, the solution of the optimization $P(\theta)$ can be regarded as a function of the hyperparameters $\theta$ \cite{jin2022solution}. Optimization techniques can be used to solve a range of energy-related problems \cite{arif2021review,khezri2022optimal,yang2018battery,drgovna2020all,esther2016survey}. Automating the formulation and solution of optimization problems is essential due to the technical skills gap, challenges in manually incorporating user preferences and constraints, and the inefficiencies in manual modifications based on user feedback. This not only conserves time but also facilitates the optimization of intricate energy systems and boosts user engagement by providing real-time feedback and insights.

The human-guided autoformalism proposed in this study is a novel approach for automatically translating natural language task specifications to optimization instances. However, this approach grapples with issues such as ambiguity, incompleteness, and user-specific preferences. Ambiguity surfaces when multiple interpretations of natural language can lead to erroneous translations of task specifications into optimization problems \cite{liu2023we}. Incompleteness poses a problem when the task description does not provide all the necessary information required for formulating the optimization problem \cite{wang2023describe}. Incorporating user-specific preferences and constraints into the optimization formulation can also be challenging \cite{wu2023tidybot}. The subsequent subsections will delineate our approach and the strategies we have devised to address the aforementioned challenges.

\subsubsection{Optimization formulation}

The first step in the optimization process is to identify the key components of the problem, such as the objective function, decision variables, and constraints, after the user provides a task description. Let [T] denote the task provided by the user (e.g., ``help to optimize my EV charging schedule to minimize cost'').  We have experimented with two approaches:
\begin{itemize}
    \item Approach 1: We inquire, ``What is the [A] in [T] formulated as an optimization problem?'', where [A] can be ``objective function'', ``decision variable'', or ``constraints''. Once these elements are identified, we request users to specify parameters in these elements, thereby translating the query into an optimization instance.
    \item Approach 2: We initially prompt the LLM with ``[T]. Identify the most important 5 parameters.''. Subsequently, we present the task [T] along with user inputs to the LLM to formulate an optimization instance.
\end{itemize}
Interestingly, the first approach tends to underperform, while the second approach typically results in the LLM accurately identifying the most relevant parameters. This discrepancy can be ascribed to the training methodology and information processing mechanisms of LLMs like GPT-4. LLMs are trained on a broad spectrum of data and not exclusively on optimization problems. Thus, when directly prompted to formulate specific components of an optimization problem, they might not yield the most accurate or helpful responses due to their generalist nature. The second approach, however, breaks down the task into smaller, digestible parts (identifying the top 5 parameters), which corresponds well with the operational mode of LLMs. LLMs excel when the tasks are simplified, and they can focus on singular concepts at a time. Moreover, the initial identification of crucial parameters provides a context to the LLM. With a grasp on the vital factors, the LLM can use this context to generate more accurate and pertinent responses while formulating the optimization problem. In contrast, without this contextual framework (in the first approach), the LLM might struggle to pinpoint and focus on the pivotal aspects of the problem. This is in line with recent approaches \cite{weichain,yao2023tree}. Furthermore, a direct request to formulate an optimization problem component might leave room for ambiguity and misinterpretation, owing to the complexity of such tasks. By identifying key parameters initially, this ambiguity can be significantly reduced or even eliminated.

Upon pinpointing the essential parameters of the optimization problem, the subsequent step involves transforming the natural language query into a computational instance using the prompt: "Write a Python code using [lib] to solve this optimization problem," where [lib] could be either CVXPY or SCIPY. This transformation translates the query into code that formats the optimization problem in a standard format interpretable by existing libraries such as CVXPY\cite{diamond2016cvxpy} or SCIPY\cite{virtanen2020scipy}. To discern the superior library, we assessed each package based on the number of accurate, error-free codes generated. Generally, we found that translating into CVXPY code yields a success rate ranging from 51\% to 80\%, whereas SCIPY showcases an impressive 71\% to 100\% success rate (refer to Fig. \ref{fig:comparison} for more details). We attribute this performance disparity to SCIPY's broader acceptance and longer existence in the Python community compared to CVXPY, which implies a greater availability of SCIPY-related content online that the model could have leveraged during training. Consequently, the LLM might have encountered more optimization problem instances solved using SCIPY than with CVXPY in its training data.\footnote{A rudimentary estimation (dated 5/29/23) reveals that a search for ``SCIPY'' on GitHub returns 7,075 repositories, whereas a search for ``CVXPY'' results in merely 230 hits. This significant difference corroborates our observations and lends credence to our preference for SCIPY for this task.}

\subsubsection{Solving an optimization and debugging}

To address the limitations of LLMs in nonlinear reasoning \cite{patel2021nlp,rae2021scaling}, we have enabled the model to interact with an external Python program execution to solve optimization tasks. A vital step in interfacing with the solver involves extracting code from the LLM's output. To accomplish this, we deploy regular expressions to isolate the relevant code block within the text. Regular expressions, as a pattern-matching language, facilitate the search for specific patterns within the text. We prompt the LLM to encapsulate the code block with a unique delimiter (like triple backticks or a phrase like ``the code starts/ends here''), which we can then search for, extracting the enclosed text. The extracted text undergoes further processing to ensure its validity and compatibility with the optimization solver. This method, we found, can extract code reliably, without necessitating a large dataset of labeled examples, unlike when training a machine learning model to identify code blocks \cite{iyer2016summarizing}.

During our development process, we encountered an important issue: the code generated by the LLM occasionally triggers errors during execution. We have classified these errors into two categories: erroneous translation into an optimization problem and syntactic bugs in an otherwise correctly translated optimization problem.\footnote{Errors can also arise due to the limitations of the optimization solvers employed to solve the generated optimization problem. Regardless of the syntactical correctness of the generated code, the solver might fail to find a feasible or optimal solution due to the problem's complexity or other constraints. However, since most energy problems involve linear or quadratic convex optimization, we presume that such errors are comparatively infrequent. We do not classify infeasibility as an error.} The first category of errors may result from ambiguity, incomplete information, or inaccurate identification of hyperparameters. The second category can often be traced back to the limitations of LLM's natural language processing capabilities in accurately capturing the specific syntax required by the solver.

\begin{figure}[t]
    \centering
    \includegraphics[width=0.4\textwidth]{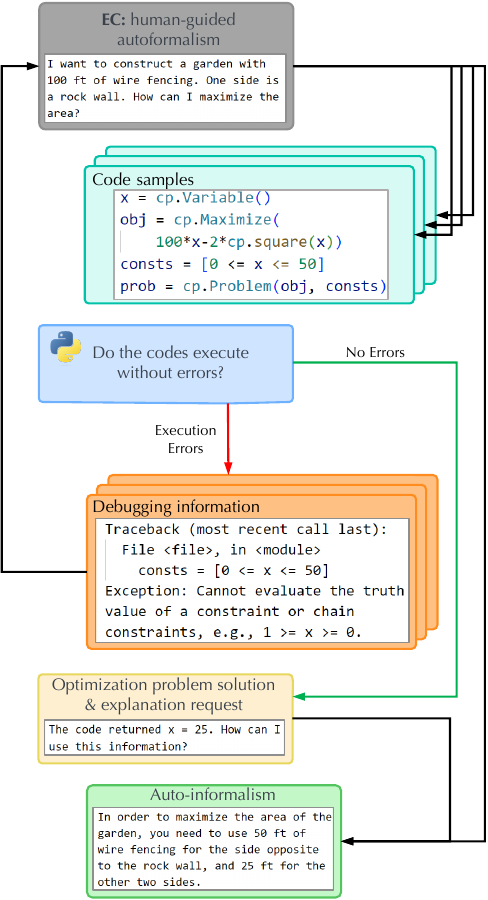}    \caption{Illustration of the process of solving the optimization problem and debugging the code generated by the LLM. The flow adheres to the \textbf{No Errors} path if the optimization code, as produced by the LLM, executes seamlessly in Python without any errors. If not, it diverts to the \textbf{Execution Errors} path. Any debugging information collected from Python at this stage is then incorporated into our refined prompt. This added context is subsequently used when requesting the LLM to generate a new code iteration.}
    \label{fig:debugging_illustration}
\end{figure}

To rectify the first type of error, which arises from inaccurate translations, our primary reliance is on user interaction and clarification of the formulated optimization (i.e., optimization auto-informalism, as detailed in the following subsection). For the second type of error, we resort to an automated process. Initially, we pinpoint the error message or unanticipated behavior identified by the solver, and this information is fed into the LLM to isolate relevant code snippets or whole functions. The LLM is then employed to generate potential remedies for the identified issue. Subsequently, we assess the proposed solutions in a new iteration to determine if they ameliorate the problem. If they fail, we persist with the process until a suitable solution emerges. This procedure is graphically represented in Fig. \ref{fig:debugging_illustration}. Our findings indicate that taking more than one sample before commencing the debugging process often results in fewer queries to the LLM (as seen in the multiple code sample blocks in Fig. \ref{fig:iteration_and_debugging}). Moreover, potential optimization codes can be tested concurrently to expedite the solution-finding process. In instances where all code results in execution errors, we can revert to the debugging process, again running it in parallel. By adhering to this procedure, we've been successful in resolving the majority of the errors encountered during our experiments.

\subsubsection{Optimization auto-informalism}
\label{sec:autoinform}
Upon discovering the optimal solution, it is the responsibility of the LLM to articulate the results in a comprehensible, natural language format. This system is designed to provide detailed explanations of the optimal solution, alongside any constraints or preferences factored in during the optimization process (refer to Steps 7a and 7b in Fig. \ref{fig:ab}). This approach, aptly termed auto-informalism, dovetails with autoformalism by offering intuitive insight into the optimization outcomes.

In order to generate an explanation of the optimization code, we prompt the LLM with the generated and revised code using a simple request: ``Please explain the following code.'' We then delve deeper, seeking clarification on the decision variables, objective function, and constraints implicated in the optimization problem.

\section{Experiments and Case Studies}
\label{sec:experiments}

In this section, we delve into a series of experiments and case studies that explore the potential of LLM-based optimization autoformalism in optimizing real-time decision-making and long-term planning for sustainability. By examining the applications of the proposed method in various domains such as EV charging, HVAC controls, and energy storage control, we aim to provide an understanding of how these technologies can improve efficiency and reduce costs. Furthermore, we will conduct cost-benefit analyses of long-term sustainable solutions, such as the installation of heat pumps and rooftop PV systems. We start with general evaluations of the proposed methods in this section, then delve into specific case studies in subsequent sections.

\subsection{General findings and analysis}
\begin{figure}[t]
    \centering
    \includegraphics[width=.7\textwidth]{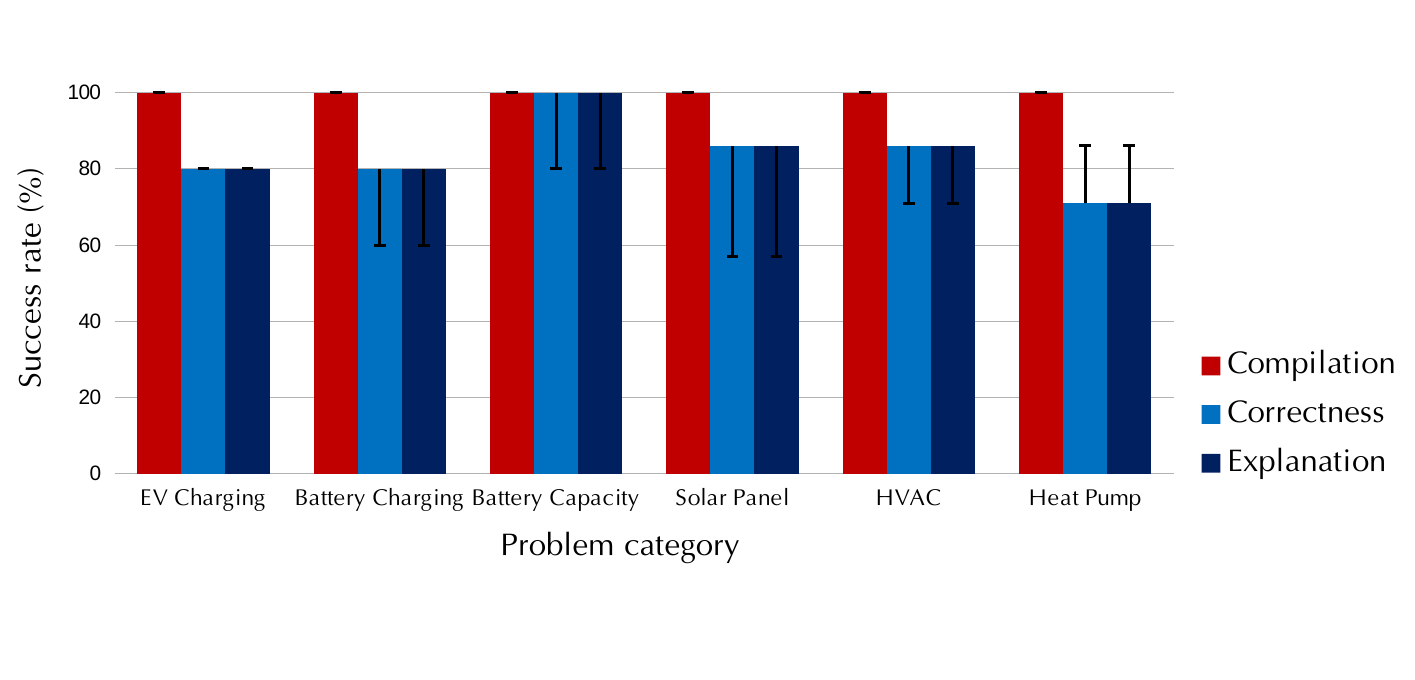}
    \caption{Success rate of EC with SCIPY in terms of the compilation rate of the generated python code, correctness of the solution found and its explanation to the user. The black lines indicate the performance change when using CVXPY.}
    \label{fig:ec_success_rate}
\end{figure}

\begin{figure}[h!]
    \centering
    \begin{subfigure}[b]{0.49\textwidth}
         \centering
         \includegraphics[width=0.99\textwidth]{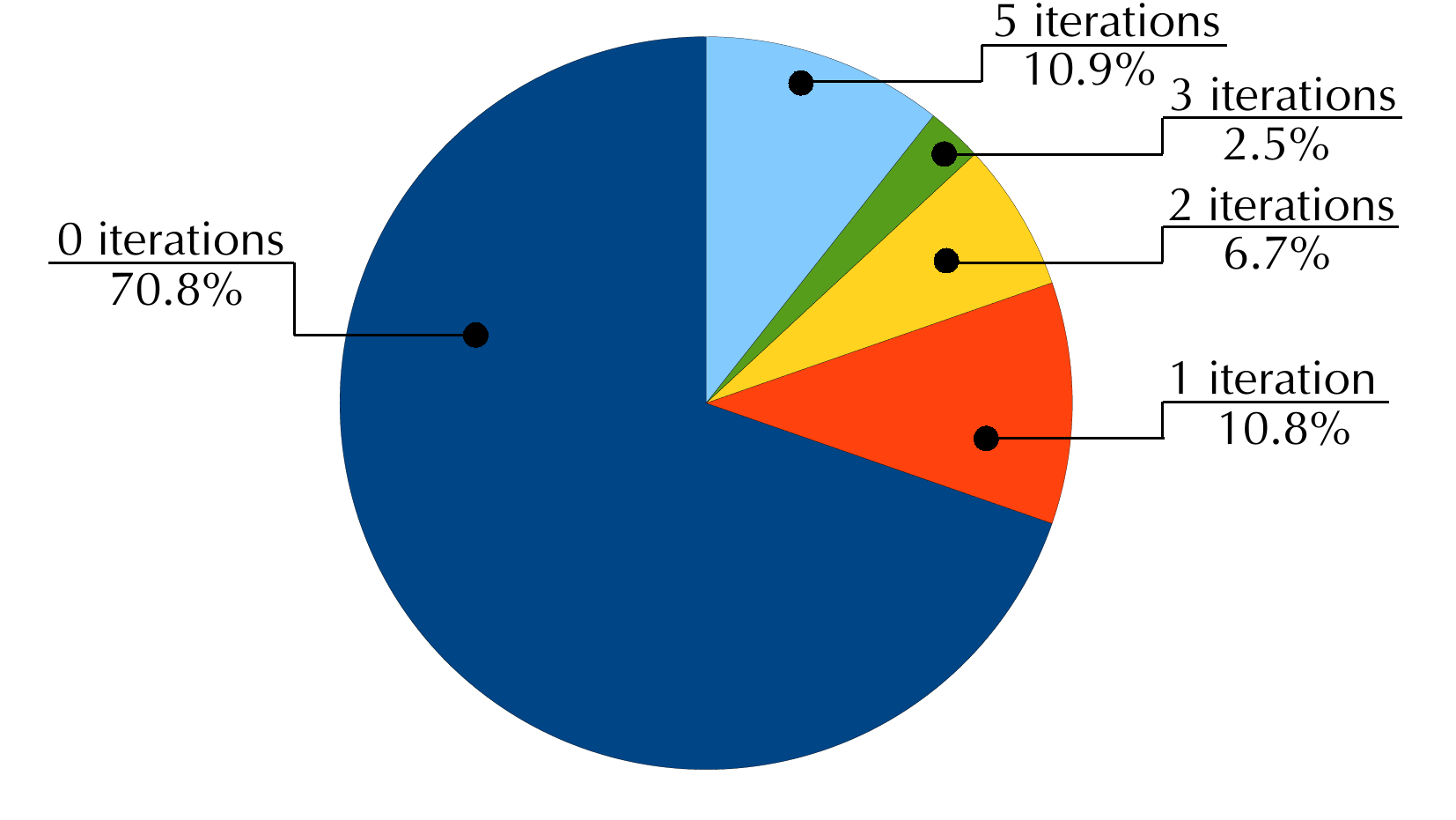}
         \caption{}
         \label{fig:y equals x}
     \end{subfigure}
    \begin{subfigure}[b]{0.49\textwidth}
         \centering
         \includegraphics[width=0.97\textwidth]{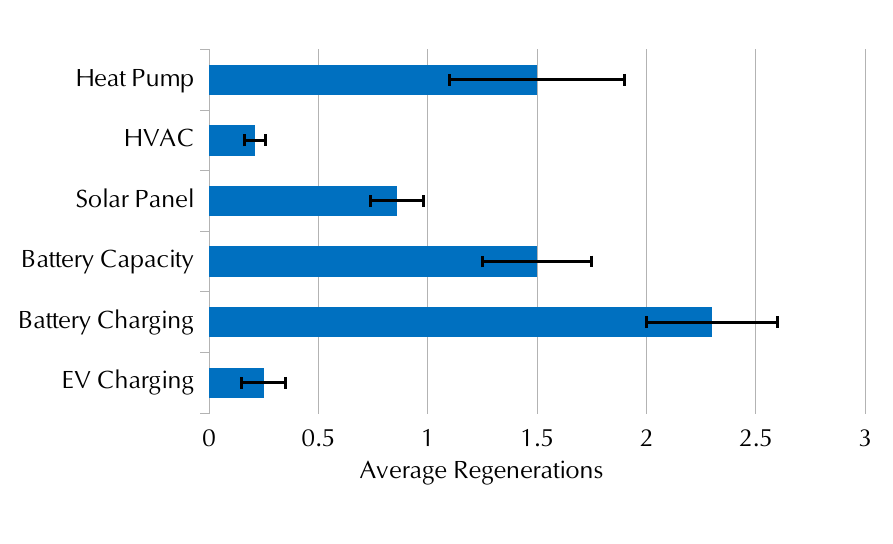}
         \caption{}
         \label{fig:y equals x}
     \end{subfigure}
    \caption{Insights on number of iterations for regeneration and debugging: (a) Comparison  of number of debugging iterations required to get an executable code for all problem categories. (b) Average number of code generations required between EC and python to get an executable code, capped at 5. The black bars show the deviation of the iterations per problem category.}
    \label{fig:iteration_and_debugging}
\end{figure}

\begin{figure}[t]
    \centering
    \includegraphics[width=.9\textwidth]{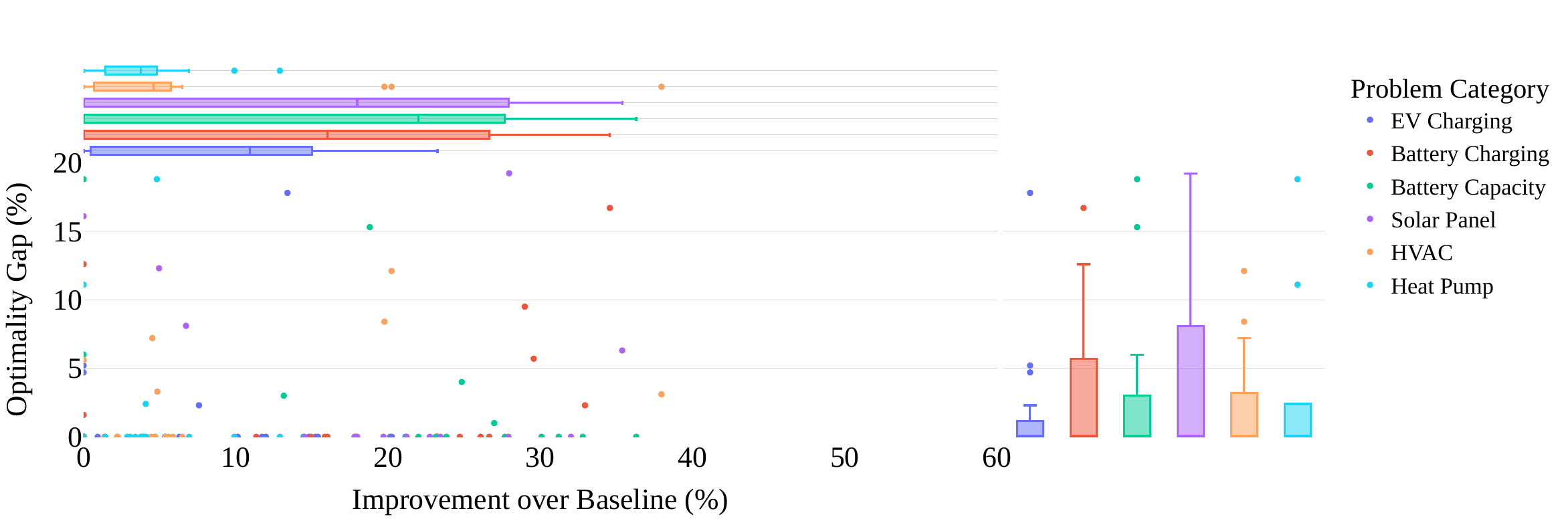}
    \caption{The average optimality gap between the exact solution of the optimization problem and the solution found by EC as well as the improvement over the baseline in terms of optimality.}
    \label{fig:optimality_gap}
\end{figure}

To elucidate the types of errors EC commits, we display the success rates concerning error-free code production, logical correctness of the code, and the ability to articulate the solution derived from the optimization code to the end user in Fig. \ref{fig:ec_success_rate}. Notably, our framework enables EC to achieve a 100\% compilation rate. Moreover, when EC generates an error-free optimization code, it consistently explains the results to the end user, indicating how to utilize them. Interestingly, SCIPY outperforms CVXPY in overall performance, which aligns with our earlier observation that SCIPY has broader recognition and use online, where most LLMs receive training. This finding implies that improving CVXPY's performance might be attainable by fine-tuning on a codebase with extensive use of the CVXPY library.

We present the average number of code regenerations needed and the average debugging iterations required to obtain syntax error-free optimization code in Fig. \ref{fig:iteration_and_debugging}. Although each subproblem within the optimization problem category exhibits different success rates---indicating that the LLMs show varying degrees of proficiency---a relatively low number of code regenerations are typically needed. In scenarios where the LLM consistently commits errors, such as incorrect usage of a function parameter name, incorporating debugging information in our framework is essential for achieving optimization solutions. When the number of generated code samples is reduced from $s=5$ to $s=1$, there is a slight increase in the required debugging iterations across all examples. This may suggest that debugging is a more efficient error-handling strategy compared to simple regeneration.

\textbf{Estimating the Probability of One-Round Autoformalism Success.} The aforementioned results allow us to infer the probability of generating valid code successfully, denoted as $p$. We postulate that the event of successfully generating valid code is independent and identically distributed for the same category of problems, following a truncated geometric distribution with a cap at 5 trials. Based on this supposition, we deduce that $\sum_{k=1}^5 k(1-p)^{k-1}p + 6\sum_{k\geq 6}(1-p)^{k-1}p = z$, where $z$ signifies the number of generations needed to achieve success (with generations capped at 5, as shown in Fig. \ref{fig:iteration_and_debugging}), and $p$ is the probability of success to be estimated. Substituting the values for $z$ from Fig. \ref{fig:iteration_and_debugging} into the equation and solving for $p$, we obtain the one-round autoformalism success rates $p=0.8, 0.25, 0.38, 0.53, 0.83, 0.38$ for the problems of EV Charging, Battery Charging, Battery Capacity, Solar Panel, HVAC Control, and Heat Pump Investment, respectively.

\textbf{Estimation of the Probability of Debugging Success.} We posit that the event of successful debugging, given the presence of an erroneous code, is independent and identically distributed across all problem classes. Using Fig. \ref{fig:iteration_and_debugging}, we first normalize the frequency of the required number of debugging iterations by the frequency of generating an erroneous code in the first run (i.e., $1-0.7=0.3$). This reveals that the frequencies of observing debugging iterations 1, 2, 3, 4, and $\geq 5$ are represented by $y_k$, where $k\in{1,2,...,5}$. Let $q$ denote the probability of successfully debugging the code. From this, we can construct a system of linear equations in the form of $(1-q)^{k-1}q = y_k$ for $k\in{1,2,3,4}$, and $1-\sum_{k=1}^4(1-q)^{k-1}q=y_5$ given that debugging is capped at 5 iterations. Using a line search between 0 and 1 in increments of 0.01, we calculate the values of $\hat{y}_1$ to $\hat{y}_5$ for each $q$ using the specified equations, and determine the mean squared error (MSE) between the vectors $\hat{y}$ and $y$. Our optimal estimate for the probability is  $q=0.26$.

\textbf{Optimality Gap and Improvement Over Baseline.} The optimality gap for a feasible solution is calculated as the ratio $v/v^* - 1$, where $v$ denotes the objective value of the candidate solution and $v^*$ signifies the optimal value of the corresponding minimization problem. In Fig. \ref{fig:ec_success_rate}, we classify test cases as correct only if they equate to the globally optimal solution value, i.e., a 0\% optimality gap. Additional results presented in Fig. \ref{fig:optimality_gap} demonstrate the average optimality gap when incorrect logic within the code results in the omission of key parameters (such as the efficiency of a battery). Our framework was evaluated using $20$ examples for each energy optimization problem, including EV charging. As depicted, the instances display an optimality gap generally within the 20\% range, with the majority under 10\%.

For the baseline method, the model was simply prompted with our question along with necessary parameters to solve the optimization problem. We noticed that even when the problem is clearly an optimization issue, LLMs may opt to generate responses by attempting to apply logic towards reaching an answer. As seen in Fig. \ref{fig:optimality_gap}, this approach does not yield favorable results in comparison to the EC framework. This can be attributed to the fact that many energy-related optimization problems of importance do not yield closed-form solutions, making the correct utilization of convex program solvers critical.

We further report on the improvement over the baseline, measured by $v_b/v-1$, where $v_b$ represents the objective value of the baseline solution. As is evident, the improvement can amount to as much as 60\%, with most instances falling within the 30\% range. Some of the problems yielding the greatest improvements include Solar Panel, Heat Pump, Battery Charging, and EV Charging, which encompass both real-time decisions and long-term investments.

\subsection{Real-time decision making}

\begin{figure}[t]
    \centering
    \includegraphics[width=0.9\textwidth]{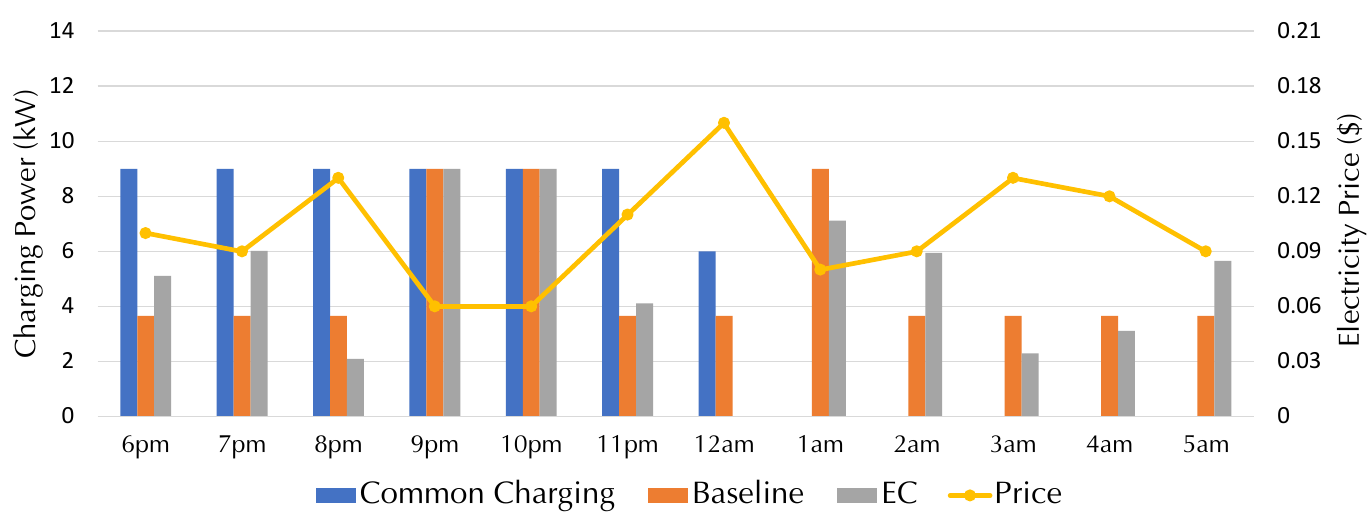}
    \caption{A comparison of EV Charging optimization to improve the cost and load on the grid. We see that EC is able to take advantage of the hours where the prices are lower compared to the baseline.}
    \label{fig:ev_charging_example}
\end{figure}

The optimal resolution of real-time decision-making problems, such as EV charging (Sec. \ref{sec:ev-charge}), HVAC controls (Sec. \ref{sec:hvac}), and energy storage control (Sec. \ref{sec:bat_control}), is paramount for enhancing energy efficiency, reducing costs, and promoting sustainable practices. These problems present distinct characteristics such as time-variability and constraints, uncertainty in demand and supply, and the necessity for adaptive decision-making in response to personalized preferences. Additionally, they often involve multiple conflicting objectives requiring balance, making them ideal candidates for optimization autoformalism. This approach facilitates the flexible integration of domain-specific knowledge with LLMs, enabling the formulation and adaptation of optimization instances through natural language interactions. Consequently, this human-guided optimization autoformalism allows stakeholders to address complex energy-related problems by customizing solutions to their unique needs and preferences, all while leveraging the extensive knowledge and reasoning capabilities of models like GPT4.

\subsubsection{Smart EV charging}
\label{sec:ev-charge}

The core issue in smart EV charging is to optimize EV charging patterns to balance power grid loads, mitigate energy costs, and fulfill user preferences \cite{arif2021review}. The usual challenges in auto-formulation of such a problem involve understanding the user's conditions (for instance, the time available for charging is contingent on the user's home charging station accessibility) and transforming these into optimization constraints, along with deft handling of temporal variables linked to the EV's charge status.

Here, we offer a typical scenario from our experiments, showing the user's initial query and the subsequent EC responses. To add context, we've appended a background processing section that discloses the code written by EC and the corresponding optimization formulation that we manually drafted for ease of understanding.

\vspace{0.5em}

\begin{mybox}
\begin{humanBox}
\noindent \humanLogo \textbf{User}: Can you help me decide what is the best way to charge my EV. 
\end{humanBox}
\begin{chatgptBox}
    
\noindent \chatgptLogo \textbf{EC}: To help you learn the best way to charge your EV, please provide the following information:
\begin{itemize}
    \item EV make and model: Knowing the specific electric vehicle you own will help determine its battery capacity, charging capabilities, and any manufacturer recommendations for charging.
    \item Type of charger: Specify whether you have access to a Level 1 (standard household outlet), Level 2 (240V outlet or dedicated charging station), or a DC fast charger (public charging station).
    \item Daily driving distance: Your average daily driving distance will help determine how often and how much you need to charge your EV to meet your transportation needs.
    \item Charging location: Please indicate if you primarily charge at home, at work, or at public charging stations. This will help determine the best charging strategy based on your access to charging infrastructure.
    \item Time of day preference for charging: Some utility companies offer lower electricity rates during off-peak hours. Let us know if you have a preferred time of day for charging or if you are open to charging during specific hours to take advantage of these rates.
\end{itemize}
\end{chatgptBox}
\begin{humanBox}
    
\noindent \humanLogo \textbf{User}: (note that the user directly answers the questions)
\begin{itemize}
    \item electric only.
    \item 15kw.
    \item I need to charge it everyday, about 70kwh.
    \item at home.
    \item I come home at 6pm and need to leave for work at 7am. 
\end{itemize}

\end{humanBox}
\begin{bgBox}
    
\noindent \bgLogo \textbf{Background processing}:
\begin{subequations} \label{eq:evcharge}
    \begin{align}
\text{minimize}_{x\in\mathbb{R}^{12}}\quad & \sum_{t=0}^{11} p_t x_t \\
\text{subject to}\quad & 0 \leq x \leq \text{charger-max},\label{eq:ev-c1}\\
& \sum_{t=0}^{11}x_t=70\label{eq:ev-c2}
\end{align}
\end{subequations} 
\begin{lstlisting}
# Code (only given with important lines)
x = cp.Variable(12)
objective = cp.Minimize(cp.sum(p * x))
constraints = [
    cp.sum(x) == 70,  # Total charge needed
    x <= charger_max,  # Charger capacity
    x >= 0,  # Non-negativity
]
problem = cp.Problem(objective, constraints)
result = problem.solve()
# Output:
Optimal charging schedule: [0.0, 0.0, 0.0, 0.0, 8.8, 8.8, 8.8, 8.8, 8.8, 8.8, 8.8, 8.8]
\end{lstlisting}
\end{bgBox}
\begin{chatgptBox}
\noindent \chatgptLogo \textbf{EC}: The output you received is the optimal charging rate schedule for your EV, which minimizes your charging cost. 
To use this information, follow these steps:
    
\begin{itemize}
    \item Plug your EV into the charger at 6 PM when you get home. Since the charging rate is close to 0 kW during the high electricity pricing period (6 PM to 10 PM), you don't need to start charging immediately.
    \item Set your EV charging to begin at 10 PM. Adjust the charging rate to 8.89 kW, which the optimization determined as the most cost-effective rate for your situation. If your charging equipment or EV does not support setting a specific charging rate, try to find the closest possible rate within the acceptable range of your equipment and vehicle.
    \item Continue charging at 8.89 kW from 10 PM until 7 AM the next day. During this time, your EV will charge at a consistent rate, taking advantage of the lower electricity pricing.
    \item By 7 AM, your EV should be fully charged (80 kWh) and ready for your daily commute.
\end{itemize}

Keep in mind that the charging rate values in the solution might not be exact due to the solver's tolerance. You may need to round or adjust the values to match the capabilities of your EV and charging equipment.
\end{chatgptBox}
\end{mybox}
\vspace{0.5em}

We observe that EC can successfully correspond user-provided information like the charging rate to the corresponding optimization parameters, such as the maximum charging rate for the constraint \eqref{eq:ev-c1} on $x_t$ for $t=0,1,...,11$. Notably, it grasps the user's charging availability based on their home and work schedules and understands the common requirement of the EV being charged before the user departs for work. This information is translated into decision variables from the time the user arrives home to the time they leave (12 $x_t$ for $t=0,1,...,11$), along with an equality constraint \eqref{eq:ev-c2} indicating the total charging should equal 70kWh as directed by the user.

The EC aptly explains its decision, ``Since the charging rate is close to 0 kW during the high electricity pricing period (6 PM to 10 PM), you don't need to start charging immediately,'' and ``During this time, your EV will charge at a consistent rate, taking advantage of the lower electricity pricing.''  Lastly, the remark that the solution may be subject to errors and is only a suggestion that needs calibration based on the capabilities of the EV and charging equipment, appropriately sets user expectations and encourages tailored adjustments.

\subsubsection{HVAC control}
\label{sec:hvac}

HVAC control is a pivotal issue that has been extensively studied \cite{ alanne2022overview}. However, the algorithms that have been developed are often more advanced than the current control panels in most buildings, leading to a disconnect between sophisticated methods and user comprehension.  The heart of the issue is to discover a setpoint that ensures a comfortable indoor climate while minimizing energy consumption. The EC system offers an intuitive interface for users, providing sensible suggestions by comprehending and factoring in their preferences in decision-making processes.

\vspace{0.5em}

\begin{mybox}
\begin{humanBox}
\noindent\humanLogo \textbf{User}: Is it possible to save money by adjusting the HVAC temperature?
\end{humanBox}

\begin{chatgptBox}
\noindent\chatgptLogo \textbf{EC}: Yes, adjusting the HVAC temperature can help save money on energy costs. To provide you with more specific recommendations, please provide the following information:
\begin{itemize}
    \item Indoor temperature setting: What is your preferred temperature setting for both heating and cooling? Knowing your current setpoint will help determine if adjustments can be made for energy savings.
    \item Occupancy schedule: What is your typical daily schedule for when people are at home and away? This information will help identify potential opportunities for adjusting the temperature when the building is unoccupied.
    \item Type of HVAC system: What type of heating and cooling system do you have (e.g., central air conditioning, heat pump, furnace, etc.)? The efficiency and potential energy savings may vary depending on the system type.
    \item Insulation and building envelope: How well insulated is your home or building? Please provide information about the insulation in the walls, attic, and windows. This will help identify any areas where improvements can be made to further reduce energy consumption.
    \item Local climate: What is the typical climate in your region (e.g., mild, cold, hot, humid, etc.)? This information will help tailor recommendations based on the specific climate conditions where you live.

\end{itemize}
\end{chatgptBox}
\begin{humanBox}
\noindent\humanLogo \textbf{User}: (note that the user directly answers the questions)\begin{itemize}
    \item preferred temperature is 65-75 
    \item full day occupancy
    \item heat pump 
    \item average  insulation
    \item hot and humid climate
\end{itemize}
\end{humanBox}

\begin{bgBox}
\noindent \bgLogo \textbf{Background processing}:
\begin{subequations}
\begin{align}
\text{minimize}_{C\in\mathbb{R}}\quad & C \\
\text{subject to}\quad & C=c\frac{|x-y|}{e}z, \label{eq:hvac-c1}\\
& h \leq x \leq k, \label{eq:hvac-c2}\\
\end{align}
\end{subequations}
where $C$, $c$, $x$, $y$, $e$, $z$, $h$, $k$ represent total cost, electricity cost per kWh, adjusted temperature, ambient temperature, heat pump efficiency, occupancy, heating setpoint and cooling setpoint.
\begin{lstlisting}
# Code (only given with important lines)
adj_temp = cp.Variable()  # HVAC adjusted indoor temperature
constraints = [
    adj_temp >= heating_setpoint,
    adj_temp <= cooling_setpoint,
]
temp_difference = cp.abs(adj_temp - ambient_temp)
energy_cost = elec_cost_per_kwh * (temp_difference / heat_pump_efficiency) * occupancy
objective = cp.Minimize(energy_cost)
problem = cp.Problem(objective, constraints)
problem.solve()

# Output:
The optimal adjusted temperature: 75.00
The minimum energy cost per day: 9.60
\end{lstlisting}

\end{bgBox}

\begin{chatgptBox}
\noindent\chatgptLogo \textbf{EC}: 
    The optimal adjusted temperature of 75°F provided by the CVXPY optimization indicates that, under the given parameters and constraints, setting your indoor temperature to 75°F would minimize your energy cost for running the heat pump.

Here's how you can use this information:
\begin{itemize}
    \item Adjust your thermostat: Set your thermostat to the optimal temperature (75$^\circ$F in this case) during hot, humid days. This will help you save energy by reducing the amount of cooling your heat pump needs to provide.
    \item Programmable thermostat: If you have a programmable thermostat, you can schedule temperature adjustments to coincide with your daily occupancy pattern. For example, you can set the thermostat to 75°F during peak usage hours and adjust it to a higher temperature when the building is unoccupied. This can help you save even more on energy costs.
    \item Monitor your energy consumption: Keep track of your energy consumption over time to verify the savings achieved by implementing these changes. Compare your energy bills before and after the adjustments to see if there's a noticeable reduction in costs.
    \item Adapt to changing conditions: Keep in mind that the optimal temperature may change depending on the outdoor temperature, humidity, and other factors. You can periodically re-run the optimization with updated parameters to find the most energy-efficient temperature settings for different conditions throughout the year.
    \item Additional energy-saving measures: While adjusting the temperature is a good first step, there are other ways to improve your home's energy efficiency. You can consider upgrading your insulation, sealing gaps and cracks in the building envelope, installing energy-efficient windows, and using shading devices like awnings or blinds to reduce solar heat gain.
    \item Remember that the optimal temperature calculated in this example is based on the specific parameters provided. To get more accurate results tailored to your situation, you may need to adjust the input values, such as electricity costs, heat pump efficiency, or local climate conditions.
\end{itemize}
\end{chatgptBox}
\end{mybox}
\vspace{0.5em}
Despite its simplicity, the optimization formulated by EC encapsulates essential constraints reflecting the user's preferences \eqref{eq:hvac-c2}, and accurately determines the relationship among critical parameters such as electricity price, adjusted/ambient temperature, and heat pump efficiency to compute the total cost \eqref{eq:hvac-c1}. Importantly, EC provides clear and actionable advice like ``Set your thermostat to the optimal temperature (75$^\circ$F in this case) during hot, humid days...'' with a clear rationale.  It also offers customized suggestions based on user conditions, along with valuable energy efficiency tips such as  ``monitor your energy consumption'', and ``adapt to changing conditions''. Nevertheless, certain limitations in the provided formulations exist. First, the cost of simplicity is the inability to perform sophisticated controls like pre-heating/cooling, which entails heating the room in anticipation of occupancy and can only be accounted for through a multiperiod formulation involving intricate room thermal dynamics \cite{naderi2022demand}. Also, the way occupancy is represented as a multiplying factor in \eqref{eq:hvac-c1} is not robust. If occupancy is zero, different temperature setpoints can appear the same, though we could still choose an energy-saving alternative. While the program does provide the correct solution due to the user specifying ``full occupancy'', the issue could become more noticeable when a multiperiod schedule is taken into account.

\subsubsection{Battery Charging Control}
\label{sec:bat_control}

This problem entails optimizing the charging and discharging cycles of a home battery system while considering a multitude of factors such as electricity pricing, solar generation, and household demand. A notable instance of this is seen in international competitions like the CityLearn Challenge \cite{citylearn}. In a winning solution \cite{khattar2023winning}, the authors drafted an optimization problem aiming to minimize electricity costs, fulfill energy demands, and adhere to battery charging constraints. Nonetheless, as the parameters were not provided by the competition environment, they suggested parameter adaptations to enhance optimization performance. This is akin to our research where selecting the appropriate optimization problem for a given context is crucial. 
Instead of deriving the context from a reward signal, as in \cite{khattar2023winning}, our context is provided by the user in the form of natural language. The EC is tasked with interpreting this context accurately to establish the right optimization problem.

\vspace{0.5em}
\begin{mybox}
\begin{humanBox}
\noindent \humanLogo \textbf{User}: I live in California. I have a battery of 20kwh. And solar panels on the roof with max 3kw electricity production. My electricity price is \$0.2 between 6pm and 10pm, \$0.1 otherwise. Assume some random electricity production and comsumption in the house for the next 24 hours. Can you write me a CVXPY code to optimize for cost by controlling the battery charging or discharging? Max charge or discharge rate of the battery is 3kw.
\end{humanBox}

\begin{bgBox}

\noindent \bgLogo \textbf{Background processing}:
\begin{subequations}
\begin{align}
\text{minimize}_{x\in\mathbb{R}^{24}}\quad & \sum_{t=1}^{24} c_t P_{e,t} \\
\text{subject to}\quad & -P \leq x_t \leq P, \quad t=1,2,\dots,24  \label{eq:bat-c1}\\
& b_0 = 0, \label{eq:bat-c2}\\
& b_t = b_{t-1} + x_t, \quad t=1,2,\dots,24\label{eq:bat-c3}\\
& 0 \leq b_t \leq B, \quad t=1,2,\dots,24 \label{eq:bat-c4}\\
& c_t = x_t - s_t + d_t, \quad t=1,2,\dots,24, \label{eq:bat-c5}
\end{align}
\end{subequations}
where $c$, $b$, $x$, $P$, $s$, $d$ represent the electricity usage, battery state, battery charging rate, maximum power of the battery, solar energy production and demand of the household, respectively.
\begin{lstlisting}
# Code (only given with important lines)
battery_capacity = 20  # kWh
max_power = 3  # kW
price_peak = 0.2  # $/kWh
price_off_peak = 0.1  # $/kWh

np.random.seed(0)
solar_production = np.random.uniform(0, 0.5, 24)
household_consumption = np.random.uniform(0, max_power, 24)
battery_power = cp.Variable(24)
battery_state = cp.Variable(24)
cost = cp.Variable(24)
objective = cp.Minimize(sum(cost))
constraints = [battery_power <= max_power,
    battery_power >= -max_power,
    battery_state >= 0,
    battery_state <= battery_capacity,
    # Battery state evolves over time
    battery_state[0] == 0,  # Assume the battery starts empty
    battery_state[1:] == battery_state[:-1] + battery_power[1:],
    battery_state[-1] == 0,
]
for i in range(24):
    constraints.append(cost[i] == (battery_power[i] - solar_production[i] + household_consumption[i]) * (price_peak if 18 <= i < 22 else price_off_peak))
prob = cp.Problem(objective, constraints)
prob.solve()
# Output:
Total cost:  1.6952066412652915
Battery power (kW):  [-3.00000056  2.2385663   1.72916783  2.15337281  2.39723956  0.91568419
  0.34981417  0.13375868  0.05146217  0.02062781  0.0104214   0.01063623
  0.0214874   0.053826    0.13999059  0.36614605  0.95844804  2.50919925
 -2.9999989  -2.99997493 -3.0000141  -3.00001957 -1.01609253 -1.04374843]
\end{lstlisting}
\end{bgBox}

\begin{chatgptBox}
\noindent \chatgptLogo \textbf{EC}: The optimization program provided generates two main outputs:
\begin{itemize}
    \item Total cost: This is the total estimated cost of your electricity consumption for the given 24 hour period, under the optimal charging and discharging policy of the battery. In this case, your total cost is estimated to be approximately \$$1.70$.
    \item This array shows the optimal charge (-) or discharge (+) rate of the battery for each hour in the 24-hour period. For example, in the first hour (-3.00000056 kW), the battery should be charged at nearly the maximum rate. In the second hour (2.2385663 kW), the battery should be discharged at roughly 2.24 kW. This indicates that the solar panels are likely generating less power than the household is consuming. This charging and discharging strategy continues throughout the 24-hour period to minimize the cost of electricity.
\end{itemize}
\end{chatgptBox}
\end{mybox}
\vspace{0.5em}
Similar to the prior case involving EV charging (Sec. \ref{sec:ev-charge}), EC successfully formulates an accurate objective that computes the weighted sum of total electricity cost, factoring in the electricity price. In the code, a for loop initiates a segment that assigns the appropriate values for each hour, constructing a 24-dimensional price vector. EC also correctly formulates the lower and upper limits for the charging and discharging variables in equation \eqref{eq:bat-c1}, the temporal interdependence of the state of charge and the charging rate in \eqref{eq:bat-c3}, and the maximum capacity constraint in \eqref{eq:bat-c4}.

The energy balance constraint \eqref{eq:bat-c5} is particularly intriguing as it exhibits LLM's innate understanding of world modeling, including this constraint even without specific user instruction. In its explanation, EC further demonstrates an understanding of physical constraints, such as ``in the first hour, the battery should be charged at nearly the maximum rate,'' and its overall objective ``to minimize the cost of electricity''.

One limitation, however, lies in the implicit assumption that excess energy produced each hour can be sold back to the grid at the same rate as it was purchased, which is not typically the case. If selling electricity back is not feasible, we could adjust the formulation by introducing a nonnegativity constraint on $c_t$ or change the objective to $\sum_{t=1}^{24}\max(c_t,0)P_{e,t}$ to zero out the compensation for surplus energy. While some assumptions about the world, like the energy balance equation \eqref{eq:bat-c5}, are valid, future work can explore methods to encourage EC to be more explicit about the assumptions it is making when constructing optimization instances.

\subsection{Long-term planning for sustainability}
\label{sec:longterm}

\subsubsection{Cost-benefit analysis of installing rooftop solar PVs}
\label{sec:pv}
To perform a cost-benefit analysis of solar PV, one must estimate the costs and benefits over the system's lifespan and compare them to a relevant alternative system \cite{thebault2021optimization}. 
Elements of this analysis encompass the initial cost of installing the solar PV system, a cost that can vary based on the system's type, size, complexity, and the availability of incentives and rebates. The operational costs, influenced by the electricity price, system efficiency, and climate conditions, also play a significant role. Generally, objectives revolve around maximizing financial savings \cite{zhu2022economically} and curtailing environmental impacts \cite{jung2021optimal}, with constraints related to installation costs, roof space, and local regulations \cite{christiaanse2021techno}. A distinctive strength of EC lies in its ability to bridge information gaps when the user neglects to provide all necessary details upfront. As a result, EC procures the needed information, potentially augmenting the user's comprehension of the matter.

\vspace{0.5em}

\begin{mybox}
\begin{humanBox}
\noindent\humanLogo \textbf{User}: Should I install solar panels on my roof? How much can it save me? 
\end{humanBox}
\begin{chatgptBox}
\noindent \chatgptLogo \textbf{EC}: Installing solar panels on your roof can be a great way to save on electricity costs and reduce your carbon footprint. However, the amount you can save depends on various factors. To help me provide an accurate estimate of the potential savings, please provide the following information:

\begin{itemize}
    \item Location: In which city or region is your home located? This will help determine the amount of sunlight your solar panels can receive.
    \item Roof size and orientation: What is the size of your roof and its orientation (e.g., facing south, east, or west)? This will affect the number of solar panels that can be installed and their efficiency.
    \item Energy consumption: What is your average monthly electricity consumption in kilowatt-hours (kWh)? This will help determine the size of the solar panel system required to meet your energy needs.
    \item Local electricity rates: What is the cost per kWh for electricity in your area? This will help calculate your potential savings from using solar power.
    \item Budget: What is your budget for installing a solar panel system? This will help determine the size and quality of the system you can install, which will affect your potential savings.

\end{itemize}
\end{chatgptBox}
\begin{humanBox}
\noindent\humanLogo \textbf{User}: (note that the user directly answers the questions)
\begin{itemize}
    \item Seattle USA
    \item 300 sq ft facing east
    \item 400 KWh per month 
    \item USD 0.13 per KWh
    \item 8000
\end{itemize}
\end{humanBox}
\begin{bgBox}

\noindent \bgLogo \textbf{Background processing}:
\begin{subequations}
\begin{align}
\text{minimize}_{U\in\mathbb{R}}\quad & U \\
\text{subject to}\quad & U=B-p\times A, \label{eq:pv-c1} \\
& R \geq A, \label{eq:pv-c2}\\
& A\times w\times 0.12 \geq M, \label{eq:pv-c3}\\
& A \geq 0, \label{eq:pv-c4}
\end{align}
\end{subequations}
where $U$, $B$, $p$, $A$, $R$, $w$, $M$ represent unused budget, budget, panel price rate, panel area, roof area, panel wattage per sqft and monthly consumption.
\begin{lstlisting}
# Code (only given with important lines)
panel_area = cp.Variable()
objective = cp.Minimize((budget - (panel_area * 10)))
constraints = [
    panel_area <= roof_area,  # Panel area should not exceed the roof area
    panel_area * panel_wattage_per_sqft * 0.12 >= monthly_consumption,  # Monthly energy production should meet or exceed consumption
    panel_area >= 0  # Panel area should be non-negative
]
problem = cp.Problem(objective, constraints)
result = problem.solve()
annual_savings = 12 * monthly_consumption * electricity_rate - (12 * (panel_area.value * panel_wattage_per_sqft * 0.12 - monthly_consumption) * electricity_rate)
# Output:
Optimal panel area (sq ft): 300.000000102021
Total cost (USD): 3000.0000010202098
Monthly energy production (kWh): 540.0000001836378
Annual savings (USD): 405.59999971352505
\end{lstlisting}

\end{bgBox}
\begin{chatgptBox}

\noindent\chatgptLogo \textbf{EC}: This information can be useful for someone considering installing solar panels on their roof. They can use the results to make an informed decision about whether or not to invest in solar panels and what size of installation would be optimal for their situation. The results provide the following insights:

\begin{itemize}
    \item Optimal panel area (sq ft): This is the recommended size of the solar panel installation that satisfies the given constraints, such as roof area and monthly energy consumption. In this case, it is nearly the maximum possible area (300 sq ft).
    \item Total cost (USD): This is the estimated cost of the solar panel installation based on the optimal panel area. In this case, the cost is \$2,999.99, which is within the budget of \$8,000.
    \item Monthly energy production (kWh): This is the estimated amount of electricity the solar panels will produce each month, based on the optimal panel area. In this case, it is 539.99 kWh, which is more than the monthly consumption of 400 kWh.
    \item Annual savings (USD): This is the estimated amount of money the solar panels will save on electricity bills over the course of a year. In this case, the annual savings are \$405.60.
\end{itemize}
Using this information, someone can determine if investing in solar panels makes financial sense for their situation. They can weigh the cost of the installation against the expected savings and consider other factors, such as environmental benefits or local incentives for renewable energy. Additionally, they can use the optimal panel area as a starting point for discussions with solar panel installers to get more accurate cost estimates and further refine their decision-making process.

\end{chatgptBox}
\end{mybox}
\vspace{0.5em}

From the above, EC shows proficiency in understanding some physical constraints, such as the stipulation that the panel area shouldn't surpass the roof area \eqref{eq:pv-c2}, and proposes that the installed area be adequate to supply the electricity demand. EC is also able to articulate all key decisions and variables—like the optimal panel area, total cost, monthly energy production, and annual savings—in a clear-cut manner. However, it does operate under a few assumptions, such as the user planning to source all electricity demand from PV \eqref{eq:pv-c3} to fully leverage the budget, and that the PV's efficiency stands at 0.12. These assumptions are generally sensible, but stating them in the explanations would make the model's workings more transparent. It is not entirely satisfactory that the model seems to underutilize the user's provided information, such as the building's location, which could potentially inform the required area-to-power conversion efficiency more accurately. Nevertheless, this is understandable given the model's lack of access to an external database. Future work that enhances this capability could be a worthwhile pursuit.

\subsubsection{Cost-benefit analysis of installing a heat pump}
\label{sec:hp}
Key considerations in the cost-benefit analysis of a heat pump include the initial cost of installation, which can vary based on the type, size, and complexity of the system, as well as the availability of incentives and rebates \cite{liu2018cost,belaid2021exploring}. The analysis also takes into account the operating costs of running a heat pump system, along with the savings compared to existing systems. 

\vspace{0.5em}

\begin{mybox}
\begin{humanBox}
\noindent \humanLogo \textbf{User}: What is the monetary advantage of installing a heat pump?
\end{humanBox}
\begin{chatgptBox}
    
\noindent \chatgptLogo \textbf{EC}: To determine the monetary advantage of installing a heat pump, please provide the following information:
\begin{itemize}
    \item Climate: What is the local climate in your area? The efficiency and potential savings of a heat pump depend on the temperature range it will operate in.
    \item Current heating and cooling system: What type of heating and cooling system do you currently have in your home? This will help determine the potential energy and cost savings from switching to a heat pump.
    \item Energy prices: What are the current prices for electricity and any other energy sources (e.g., natural gas, oil, propane) used in your home? This will be used to compare the operating costs of different systems.
    \item Heat pump efficiency: What is the efficiency rating (e.g., Seasonal Energy Efficiency Ratio or SEER, and Heating Seasonal Performance Factor or HSPF) of the heat pump you are considering? Higher efficiency ratings indicate a more energy-efficient system.
    \item Installation and maintenance costs: What are the estimated costs of purchasing, installing, and maintaining the heat pump? These costs will factor into the overall monetary advantage of the system.
\end{itemize}
Once you provide this information, we can help you estimate the potential monetary advantages of installing a heat pump.
\\
\end{chatgptBox}
\begin{humanBox}
    
\noindent \humanLogo \textbf{User}: (note that the user directly answers the questions)
\begin{itemize}
    \item Mild
    \item Central air conditioning
    \item 0.75 per KWh
    \item I don't know the rating 
    \item 200 per year

\end{itemize}
\end{humanBox}
\begin{bgBox}
\noindent \bgLogo \textbf{Background processing}:
\begin{subequations}
\begin{align}
\text{minimize}_{C\in\mathbb{R}}\quad & C \\
\text{subject to}\quad & C=a-b,  \label{eq:hp-c1}\\
& b \geq 0, \label{eq:hp-c2}\\
& a \geq 0, \label{eq:hp-c3}\\
& b = c \times 0.75, \label{eq:hp-c4}\\
& a = d \times 0.75 + 200, \label{eq:hp-c5}
\end{align}
\end{subequations}
where $C$, $a$, $b$, $c$, $d$ represent total cost, heat pump cost, central AC cost, central AC energy consumption and heat pump energy consumption.
\begin{lstlisting}
# Code (only given with important lines)
central_AC_cost = cp.Variable()
heat_pump_cost = cp.Variable()
constraints = [
    central_AC_cost >= 0,
    heat_pump_cost >= 0,
    central_AC_cost == central_AC_energy_consumption * 0.75,
    heat_pump_cost == heat_pump_energy_consumption * 0.75 + 200,
]
problem = cp.Problem(cp.Minimize(heat_pump_cost - central_AC_cost), constraints)
problem.solve()
# Output:
Annual savings with heat pump: $550.00
#end
\end{lstlisting}

\end{bgBox}

\begin{chatgptBox}
\noindent \chatgptLogo \textbf{EC}: With the calculated annual savings of \$550.00, you can now use this information to make an informed decision about whether or not to install a heat pump in your home. Here's how you can use this information:
\begin{itemize}
    \item Evaluate the investment payback period: Compare the upfront cost of purchasing and installing the heat pump to the annual savings. By dividing the total upfront cost by the annual savings, you can determine the number of years it will take for the heat pump to pay for itself through energy savings. For example, if the heat pump costs \$5,000 to purchase and install, the payback period would be about 9.1 years (\$5,000 / \$550).
    \item Compare with other options: Use the annual savings to compare the heat pump with alternative heating and cooling systems. Research the energy efficiency, operating costs, and maintenance costs of other systems (such as furnaces, boilers, or air source heat pumps) to determine if the heat pump is the most cost-effective and environmentally friendly option for your needs.
    \item Assess your long-term plans: Consider how long you plan to stay in your current home. If the payback period is longer than the time you plan to live in the house, it may not make financial sense to install the heat pump. However, if you expect to live in the house for a long time, the heat pump could provide significant long-term savings.
    \item Evaluate environmental impact: By calculating the difference in energy consumption between the central air conditioning system and the heat pump, you can estimate the environmental impact of each system. Installing a heat pump can help reduce your home's carbon footprint, contributing to a more sustainable future.
    \item Apply for incentives or rebates: Some governments or utilities offer financial incentives for installing energy-efficient heating and cooling systems like heat pumps. Research local and national rebate programs to see if you qualify for any financial assistance, which could further reduce the upfront cost and shorten the payback period.
    \item Consult with a professional: Before making a decision, consult with a heating, ventilation, and air conditioning (HVAC) professional who can assess your home's specific needs, recommend the best heat pump model, and provide accurate cost and savings estimates.
\end{itemize}
By analyzing this information and considering these factors, you can make an educated decision about whether installing a heat pump is the right choice for your home.
\end{chatgptBox}
\end{mybox}
\vspace{0.5em}

In a manner akin to the PV installation scenario in Sec. \ref{sec:pv}, EC proficiently identifies some pivotal relationships, including the annual operating cost of a central AC \eqref{eq:hp-c4} and a heat pump \eqref{eq:hp-c5}. However, it doesn't fully illuminate the interpretability of some vital parameters employed in the optimization. Notably, EC's approach to cost-benefit analysis zeroes in on optimizing annual savings, leaving the actual estimation of the payback period to the user. For instance, EC calculates that the annual savings with a heat pump equals \$550, then elaborates on how to use this data for investment decisions, suggesting ``If the purchase and installation of the heat pump cost \$5,000, the payback period would equate to around 9.1 years (\$5,000 / \$550).'' Considering the user did not provide any information regarding the initial cost of purchasing and installing the heat pump, this strategy appears logical. Beyond providing users with the calculation, EC also suggests users to compare alternatives, plan long-term strategies, evaluate environmental impact, and seek incentives or rebates. Lastly, it advises users to consult with a professional. This showcases EC's ability to not just provide a decision, but also offer the context and information necessary to empower users to make more cost-effective and environmentally conscious choices.

\subsubsection{Battery sizing optimization for long-term electricity bill reduction}

Figuring out the optimal sizing of a home battery system to minimize long-term electricity costs calls for a balance between the initial capital outlay and anticipated savings, taking into account variables like demand and the price of electricity \cite{hesse2017economic,akram2018optimal}. Here, our objective is not to arrive at a professionally precise computation but to provide insight to users about the diverse factors playing a role in such decision-making processes.

\vspace{0.5em}

\begin{mybox}
\begin{humanBox}
\noindent \humanLogo \textbf{User}: I live in a sunny location with some sunlight throughout the day. However, I cannot sell excess energy to the grid. I want to buy a battery to sustain my electricity usage in the evenings which is 30 kWh every day. Price of the batteries increases with the battery sizes' square. I want to learn what would be the best battery size to reduce my total cost including the battery cost and buying electricity from the grid.
\end{humanBox}
\begin{chatgptBox}
    
\noindent \chatgptLogo \textbf{EC}: \begin{itemize}
    \item Battery Cost: Could you provide a function or a method to calculate the cost of the battery based on its capacity? As you mentioned that the price of batteries increases with the battery size's square, we'd need the specifics to come up with a calculation.
    \item Efficiency of the battery system: This is important because not every kilowatt-hour (kWh) of electricity that you put into the battery will be available for use – some will be lost in the charging and discharging process.
    \item Amount of Solar Generation: How much solar energy (in kWh) are you able to generate per day? This will be critical to understand how much excess energy you can store in your battery system during the day.
    \item Electricity Rate: How much are you currently paying per kWh for electricity from the grid? This will help in comparing the cost of grid electricity with the cost of storing and using solar energy.
    \item Battery Lifespan: How many years does the battery last? This will be important for calculating the overall cost over the battery's lifespan, allowing us to work out a cost per kWh over the battery's lifetime.
\end{itemize}
\end{chatgptBox}
\begin{humanBox}
    
\noindent \humanLogo \textbf{User}: (note that the user directly answers the questions)
\begin{itemize}
    \item Battery cost is the price of the battery times the square of battery capacity, where the price of the battery is 10 dollars per kilowatt hour.
    \item 0.95
    \item It produces 10kwh each day
    \item \$$0.25$
    \item 2 years
\end{itemize}
\end{humanBox}

\begin{bgBox}
\noindent \bgLogo \textbf{Background processing}:
    
\begin{subequations}
\begin{align}
\text{minimize}_{B\in\mathbb{R}}\quad & P_{\text{bat}} B^2+(P_e \times 365 \times Y)G\label{eq:bat-c1} \\
\text{subject to} \quad &G \geq D - S - B * \eta_{\text{bat}}  \label{eq:bat-c2} \\
&G \geq 0,\label{eq:bat-c3}
\end{align}
\end{subequations}
where $P_{\text{bat}}$, $B$, $P_e$, $Y$, $G$, $D$, $S$, $\eta_{\text{bat}}$ are unit cost of the battery, battery size, price of electricity from the grid, number of years to consider, electricity usage from grid each day, demand in the evenings, solar energy production in the morning and the efficiency of the battery, respectively.

\begin{lstlisting}
# Code (only given with important lines)
battery_capacity = cp.Variable(1)
battery_cost = price_battery * battery_capacity**2
grid_electricity_cost = electricity_rate * cp.maximum(30 - battery_efficiency * battery_capacity - solar_generation, 0)
total_cost = battery_cost + 365 * battery_lifespan * grid_electricity_cost
problem = cp.Problem(cp.Minimize(total_cost))
problem.solve()
# Output:
Optimal Battery Size: 9.12505953102464 kWh
\end{lstlisting}
\end{bgBox}

\begin{chatgptBox}
\noindent \chatgptLogo \textbf{EC}: The optimal battery size of approximately 9.12 kWh means that this is the most cost-effective battery size for your specific situation, based on the parameters you provided.

Here's how you can use this information:
\begin{itemize}
    \item Purchasing a battery: When shopping for a home battery storage system, look for one that offers a usable capacity of around 9.12 kWh. If the exact size isn't available, you should consider the ones closest to this capacity.
    \item Understanding your energy use: This calculation shows that a 9.12 kWh battery should store enough solar energy during the day to cover your typical electricity usage in the evenings.
    \item Maximizing your savings: By using a 9.12 kWh battery, you should be able to minimize your reliance on the grid, thereby reducing your electricity bills over the two-year lifespan of the battery.
    \item Evaluating your options: While this analysis provides a theoretical optimum, other practical considerations may apply when selecting a battery system. You may want to consider other factors such as warranty period, brand reputation, physical size, maintenance requirements, and potential for future expansion of your solar system.
\end{itemize}

\end{chatgptBox}
\end{mybox}
\vspace{0.5em}

Firstly, EC accurately identifies the objective \eqref{eq:bat-c1}, encompassing the initial purchase and installation costs and the yearly electricity expenses over the chosen period. It also successfully encodes the appropriate energy balance in \eqref{eq:bat-c2}. Notably, in this instance, all the details offered by the users are wholly utilized in the formulation.  One potential limitation is that the optimal battery size is lower than the total solar production, which could lead to under-utilization of some solar energy. However, this would require a more granular, hourly formulation to address, which might not always have the necessary data available. The explanations that EC provides are thorough and balanced, and should aid users in their decision-making.

\section{Discussions and Conclusion}
\label{sec:conclude}

While the long-term aspiration of employing conversational AI for sustainability has seemed out of reach for a considerable time, we demonstrate how large language models offer new pathways towards this objective. Importantly, by providing a natural and intuitive conversational interface, the EC framework assists individuals in engaging more consciously and thoughtfully about their energy use. As individuals grow more aware of their energy usage patterns, they are more likely to adopt sustainable practices, such as responsible decision-making regarding energy consumption, utilization of energy-efficient appliances, or installation of solar panels.

In the scenario of EV charging, a typical household, using most appliances, consumes about 3.5 kWh of electricity during peak hours \cite{peakpower}. In contrast, an average EV charger has a capacity of around 10 kWh. Consequently, charging an EV during peak periods can triple a household's electrical load, placing significant stress on the existing power infrastructure. Power distribution companies, aiming for balanced daily electricity usage, generally offer lower rates during off-peak hours, motivating users to alter their EV charging schedules. However, research indicates that factors beyond financial incentives can influence changes in user behavior towards energy efficiency \cite{stern1986effectiveness,ratliff2014social,asensio2015nonprice,wang2022more}, a phenomenon known as the ``energy-efficiency gap'' \cite{gerarden2017assessing}. The EC framework aids in reducing this barrier, simplifying the process to a conversation with the system. Consider an off-peak rate of \$0.06/kWh versus an on-peak rate of \$0.14/kWh \cite{rate1}. In 2022, with 3 million EVs on the road \cite{EVestimate1}, adopting off-peak charging habits every day can lead to an annual saving of \$876 million. As the adoption of EVs continues to grow, these savings could escalate to an impressive \$21.3 billion by 2035 with 73 million EVs \cite{EVestimate2}. Charging EVs during off-peak hours not only benefits consumers financially but also lessens the grid load, averting potential grid failure, stabilizing electricity prices, and possibly curtailing the need for new power plants.

Despite its significant promise, a critical limitation of the EC framework lies in its dependence on LLMs' capability to accurately formulate optimization problems, ensuring proper alignment of objectives and constraints. Incorrectly formulated problems can lead to ineffective or harmful solutions, which is particularly concerning considering these systems' perceived reliability. Auto-informalism (Sec. \ref{sec:autoinform}), as illustrated in our case studies, presents a viable means of counteracting this issue. By enhancing auto-informalism's effectiveness, we can scrutinize problem formulation more carefully, potentially alerting users to any detected issues or implicit assumptions made during the process. Additional methods to enhance machine learning model reliability, such as incorporating validation routines, bolstering robustness, and expanding user feedback loops, can further fortify the EC framework.

While this paper merely sketches out the potential, the proposed shift towards human-guided optimization autoformalism could disrupt industries, democratize access to sophisticated technologies, and set the stage for a more equitable and sustainable future. Ultimately,
the unlocking of this technology's potential hinges on continual refinement and maintaining cognizance of its limitations during the development and deployment.

\bibliographystyle{alpha}
\bibliography{refs}

\end{document}